\documentclass[letterpaper, 10 pt, conference]{ieeeconf}  

\IEEEoverridecommandlockouts                              
\usepackage{amsmath} 
\usepackage{algorithmic}
\usepackage{algorithm}
\usepackage{amssymb}  
\usepackage{url}
\usepackage{multirow}
\usepackage{graphicx}
\usepackage{caption}
\DeclareMathOperator{\tr}{tr}

\title{\LARGE \bf
Random Subspace Two-dimensional LDA for Face Recognition*
}

\author{Garrett Bingham$^{1}$
\thanks{*This work was supported by the National Science Foundation (NSF) under DMS Grant Number 1659288.}
\thanks{$^{1}$G. Bingham is an undergraduate majoring in Computer Science \& Mathematics at Yale University.
{\tt\small garrett.bingham at yale.edu}}%
}
\begin{document}

\maketitle
\thispagestyle{plain} 
\pagestyle{plain}

\begin{abstract}

In this paper, a novel technique named random subspace two-dimensional LDA (RS-2DLDA) is developed for face recognition.  This approach offers a number of improvements over the random subspace two-dimensional PCA (RS-2DPCA) framework introduced by Nguyen et al. \cite{rs2dpca}.  Firstly, the eigenvectors from 2DLDA have more discriminative power than those from 2DPCA, resulting in higher accuracy for the RS-2DLDA method over RS-2DPCA.  Various distance metrics are evaluated, and a weighting scheme is developed to further boost accuracy.  A series of experiments on the MORPH-II and ORL datasets are conducted to demonstrate the effectiveness of this approach.  

\end{abstract}


\section{INTRODUCTION}

Face recognition has numerous applications in surveillance and authentication systems, yet it remains a difficult problem.  Faces of different people can appear very similar, while images of one person are often quite different.  Many approaches have been tested for face recognition.  In recent years, two-dimensional variants of well-known feature extraction methods such as principal component analysis (PCA) and linear discriminant analysis (LDA) have received growing attention.  They generally achieve higher accuracy and are computationally efficient because they require fewer coefficients for image representation.  Encouraged by the work in \cite{rs2dpca}, in which the random subspace method is applied to two-dimensional PCA, we propose a new algorithm and evaluate its performance on the MORPH-II and ORL datasets.  

The remainder of this paper is organized as follows: Section 2 gives a summary of 2DPCA and its variants; Section 3 does the same for 2DLDA; in Section 4 we review the random subspace method and its previous application to 2DPCA; Section 5 is dedicated to our new approach, RS-2DLDA; experiments are conducted in Section 6; we conclude in Section 7.

\section{TWO-DIMENSIONAL PRINCIPAL COMPONENT ANALYSIS}

\subsection{Introduction}

Principal component analysis (PCA) is a widely used feature extraction and dimension reduction technique.  In PCA-based face recognition, two-dimensional image matrices must first be transformed into one-dimensional vectors.  The vectorized images are usually of high dimension.  This makes it difficult to calculate the covariance matrix accurately when there is a relatively small number of training samples.  In \cite{2dpca}, Yang et al. proposed two-dimensional PCA (2DPCA), which differs from PCA in that the image matrices are not transformed into vectors.  Instead, an image covariance matrix is constructed from the original image matrices.  The main advantage of 2DPCA over PCA is that the size of the image covariance matrix is much smaller.  As a result, it is easier to evaluate it accurately, and it is less computationally expensive to determine the corresponding eigenvectors.

\subsection{Right 2DPCA}

Right 2DPCA (R2DPCA) is simply 2DPCA as originally proposed by Yang et al. in \cite{2dpca}.  We refer to it as Right 2DPCA to distinguish it from other generalizations of 2DPCA.\\ 

\subsubsection{Algorithm}

Let $\mathbf{A}$ be a collection of $M$ image matrices of dimension $m \times n$, where $\mathbf{A}_j$ represents the $j$th image matrix, $j = 1, \ldots, M$.  We wish to project each image matrix $\mathbf{A}_j$ onto an $n$-dimensional vector $\mathbf{V}$, resulting in an $m$-dimensional projected feature vector $\mathbf{Y}_j$.  
\begin{equation}
\mathbf{Y}_j = \mathbf{A}_j\mathbf{V}.
\end{equation}
We choose $\mathbf{V}$ such that the scatter of all projected feature vectors $\mathbf{Y}$ is maximized.  Equivalently, we seek $\mathbf{V}$ that maximizes the trace of the covariance matrix of the projected feature vectors $\mathbf{S}_v$.  Thus, we wish to maximize
\begin{equation} \label{eq:J(V)}
J(\mathbf{V}) = \tr (\mathbf{S}_v),
\end{equation}
where $\tr (\mathbf{S}_v)$ denotes the trace of $\mathbf{S}_v$.
\begin{align}
\mathbf{S}_v &= E[(\mathbf{Y} - E[\mathbf{Y}])(\mathbf{Y} - E[\mathbf{Y}])]^T\\
&= E[(\mathbf{AV} - E[\mathbf{AV}])(\mathbf{AV} - E[\mathbf{AV}])]^T\\
&= E[(\mathbf{A} - E[\mathbf{A}])\mathbf{V}][(\mathbf{A} - E[\mathbf{A}])\mathbf{V}]^T
\end{align}
therefore,
\begin{equation}
\tr(\mathbf{S}_v) = \mathbf{V}^T[E[(\mathbf{A} - E[\mathbf{A}])^T(\mathbf{A} - E[\mathbf{A}])]]\mathbf{V}.
\end{equation}
Define the matrix $\mathbf{G}_r$ as
\begin{equation}
\mathbf{G}_r = E[(\mathbf{A} - E[\mathbf{A}])^T(\mathbf{A} - E[\mathbf{A}])].
\end{equation}
From its definition, we know that $\mathbf{G}_r$ is an $n \times n$ positive semi-definite matrix.  It can be evaluated directly using the training image matrices.  Let the average image of all training images be denoted by $\mathbf{\bar{A}}$, so that
\begin{equation} \label{A_bar}
\mathbf{\bar{A}} = \frac{1}{M}\sum_{j=1}^M \mathbf{A}_j.
\end{equation}
We can then approximate $\mathbf{G}_r$ by $\mathbf{\hat{G}}_r$ where
\begin{equation}
\mathbf{\hat{G}}_r = \frac{1}{M} \sum_{j=1}^M (\mathbf{A}_j - \mathbf{\bar{A}})^T(\mathbf{A}_j - \mathbf{\bar{A}}).
\end{equation}
Equation (\ref{eq:J(V)}) can instead be expressed as
\begin{equation} \label{J(V)_2}
J(\mathbf{V}) = \mathbf{V}^T\mathbf{\hat{G}}_r\mathbf{V}.
\end{equation}
It has been shown that the vector which maximizes (\ref{J(V)_2}) is the eigenvector of $\mathbf{\hat{G}}_r$ corresponding to the largest eigenvalue.  In general it is not enough to select just one vector for projection.  Normally an orthonormal set of vectors $\mathbf{V}_1, \ldots, \mathbf{V}_d$ are chosen.  These are the eigenvectors of $\mathbf{\hat{G}}_r$ corresponding to the $d$ largest eigenvalues.\\  

\subsubsection{Feature Extraction}

The optimal projection vectors can be used for feature extraction.  Let $\mathbf{V} = [\mathbf{V}_1, \ldots, \mathbf{V}_d]$ be an $n \times d$ matrix.  Then for a given image $\mathbf{A}_j$, let
\begin{equation} \label{eq:feature_extraction}
\mathbf{Y}_{j,k} = \mathbf{A}_j\mathbf{V}_k, \qquad k = 1, \ldots, d.
\end{equation}
The set of projected feature vectors $\mathbf{Y}_{j,1}, \ldots, \mathbf{Y}_{j,d}$ are called the principal component vectors of the image $\mathbf{A}_j$.  In 2DPCA the principal components are vectors, not scalars.  The principal component vectors can be used to form an $m \times d$ feature matrix $\mathbf{Y}_j = [\mathbf{Y}_{j,1}, \ldots, \mathbf{Y}_{j,d}]$.\\

\subsubsection{Classification}

Given two arbitrary feature matrices $\mathbf{Y}_i$ and $\mathbf{Y}_j$, the distance between them can be calculated using the Frobenius norm $\|\mathbf{Y}_i - \mathbf{Y}_j\|_F$.  Other norms can also be considered.  Once all pairwise distances between feature matrices have been calculated, a $k$-nearest neighbor (KNN) algorithm is used for classification.  \\

\subsubsection{Image Reconstruction}

Since $\mathbf{V}_1, \ldots, \mathbf{V}_d$ are orthonormal, from (\ref{eq:feature_extraction}) we can obtain a reconstruction of image $\mathbf{A}_j$:
\begin{equation}
\mathbf{\tilde{A}}_j = \mathbf{Y}_j\mathbf{V}^T = \sum_{k=1}^d \mathbf{Y}_{j, k}\mathbf{V}_k^T.
\end{equation}
$\mathbf{\tilde{A}}_j$ is of the same dimension as $\mathbf{A}_j$.  If $d = n$, then $\mathbf{\tilde{A}}_j = \mathbf{A}_j$.  Otherwise, $\mathbf{\tilde{A}}_j$ is an approximation for $\mathbf{A}_j$.  See Fig. \ref{fig:orl_reconstructR}.

\subsection{Left 2DPCA}

Hong et al. showed in \cite{g2dpca} that 2DPCA is equivalent to PCA if each row of an image matrix is considered as a computational unit.  A natural extension would then be to consider each column of an image matrix as a computational unit.  This is called Left 2DPCA (L2DPCA) because the images are projected by a left matrix multiplication as opposed to a right matrix multiplication in conventional 2DPCA.  The algorithm is formulated in \cite{2d2pca}.  It is important to consider both Right and Left 2DPCA because the rows of an image may contain vital discriminatory information that is lacking in the columns, and vice-versa.  The algorithm for L2DPCA largely mimics that of R2DPCA, with a few small changes.  \\

\subsubsection{Algorithm}

Let $\mathbf{A}$ be a collection of $M$ image matrices of dimension $m \times n$, where $\mathbf{A}_j$ represents the $j$th image matrix, $j = 1, \ldots, M$.  We wish to project each image matrix $\mathbf{A}_j$ onto an $m$-dimensional vector $\mathbf{U}$, resulting in an $n$-dimensional projected feature vector $\mathbf{Y}_j$.  
\begin{equation}
\mathbf{Y}_j = \mathbf{U}^T\mathbf{A}_j
\end{equation}
We choose $\mathbf{U}$ such that the scatter of all projected feature vectors $\mathbf{Y}$ is maximized.  Equivalently, we seek $\mathbf{U}$ that maximizes the trace of the covariance matrix of the projected feature vectors $\mathbf{S}_u$.  Thus, we wish to maximize
\begin{equation} \label{eq:J(U)}
J(\mathbf{U}) = \tr (\mathbf{S}_u),
\end{equation}
where $\tr (\mathbf{S}_u)$ denotes the trace of $\mathbf{S}_u$.
\begin{align}
\mathbf{S}_u &= E[(\mathbf{Y} - E[\mathbf{Y}])^T(\mathbf{Y} - E[\mathbf{Y}])]\\
&= E[(\mathbf{U}^T\mathbf{A} - E[\mathbf{U}^T\mathbf{A}])^T(\mathbf{U}^T\mathbf{A} - E[\mathbf{U}^T\mathbf{A}])]\\
&= E[[\mathbf{U}^T(\mathbf{A} - E[\mathbf{A}])]^T[\mathbf{U}^T(\mathbf{A} - E[\mathbf{A}])]]
\end{align}
therefore,
\begin{equation}
\tr(\mathbf{S}_u) = \mathbf{U}^T[E[(\mathbf{A} - E[\mathbf{A}])(\mathbf{A} - E[\mathbf{A}])^T]]\mathbf{U}.
\end{equation}
Define the matrix $\mathbf{G}_l$ as
\begin{equation}
\mathbf{G}_l = E[(\mathbf{A} - E[\mathbf{A}])(\mathbf{A} - E[\mathbf{A}])^T].
\end{equation}
From its definition, we know that $\mathbf{G}_l$ is an $m \times m$ nonnegative definite matrix.  It can be evaluated directly using the training image matrices.  Let the average image of all training images be denoted by $\mathbf{\bar{A}}$ as defined in (\ref{A_bar}).  We can then approximate $\mathbf{G}_l$ by $\mathbf{\hat{G}}_l$, where 
\begin{equation}
\mathbf{\hat{G}}_l = \frac{1}{M} \sum_{j=1}^M (\mathbf{A}_j - \mathbf{\bar{A}})(\mathbf{A}_j - \mathbf{\bar{A}})^T.
\end{equation}
Equation (\ref{eq:J(U)}) can instead be expressed as
\begin{equation} \label{J(U)_2}
J(\mathbf{U}) = \mathbf{U}^T\mathbf{\hat{G}}_l\mathbf{U}.
\end{equation}
It has been shown that the vector which maximizes (\ref{J(U)_2}) is the eigenvector of $\mathbf{\hat{G}}_l$ corresponding to the largest eigenvalue.  In general it is not enough to select just one vector for projection.  Normally an orthonormal set of vectors $\mathbf{U}_1, \ldots, \mathbf{U}_d$ are chosen.  These are the eigenvectors of $\mathbf{\hat{G}}_l$ corresponding to the $d$ largest eigenvalues.\\  

\subsubsection{Feature Extraction}

The optimal projection vectors can be used for feature extraction.  Let $\mathbf{U} = [\mathbf{U}_1, \ldots, \mathbf{U}_d]$ be an $m \times d$ matrix.  Then for a given image $\mathbf{A}_j$, let
\begin{equation} \label{eq:feature_extraction2}
\mathbf{Y}_{j,k} = \mathbf{U}_k^T\mathbf{A}_j, \qquad k = 1, \ldots, d.
\end{equation}
The set of projected feature vectors $\mathbf{Y}_{j,1}, \ldots, \mathbf{Y}_{j,d}$ are called the principal component vectors of the image $\mathbf{A}_j$.  In 2DPCA the principal components are vectors, not scalars.  The principal component vectors can be used to form a $d \times n$ feature matrix $\mathbf{Y}_j = [\mathbf{Y}_{j,1}^T, \ldots, \mathbf{Y}_{j,d}^T]^T$.\\

\subsubsection{Classification}

Classification here is equivalent to that in R2DPCA.

\subsubsection{Image Reconstruction}

\begin{figure*}
\centering
\begin{minipage}{0.5\textwidth}
\centering
\includegraphics[width=\textwidth,keepaspectratio]{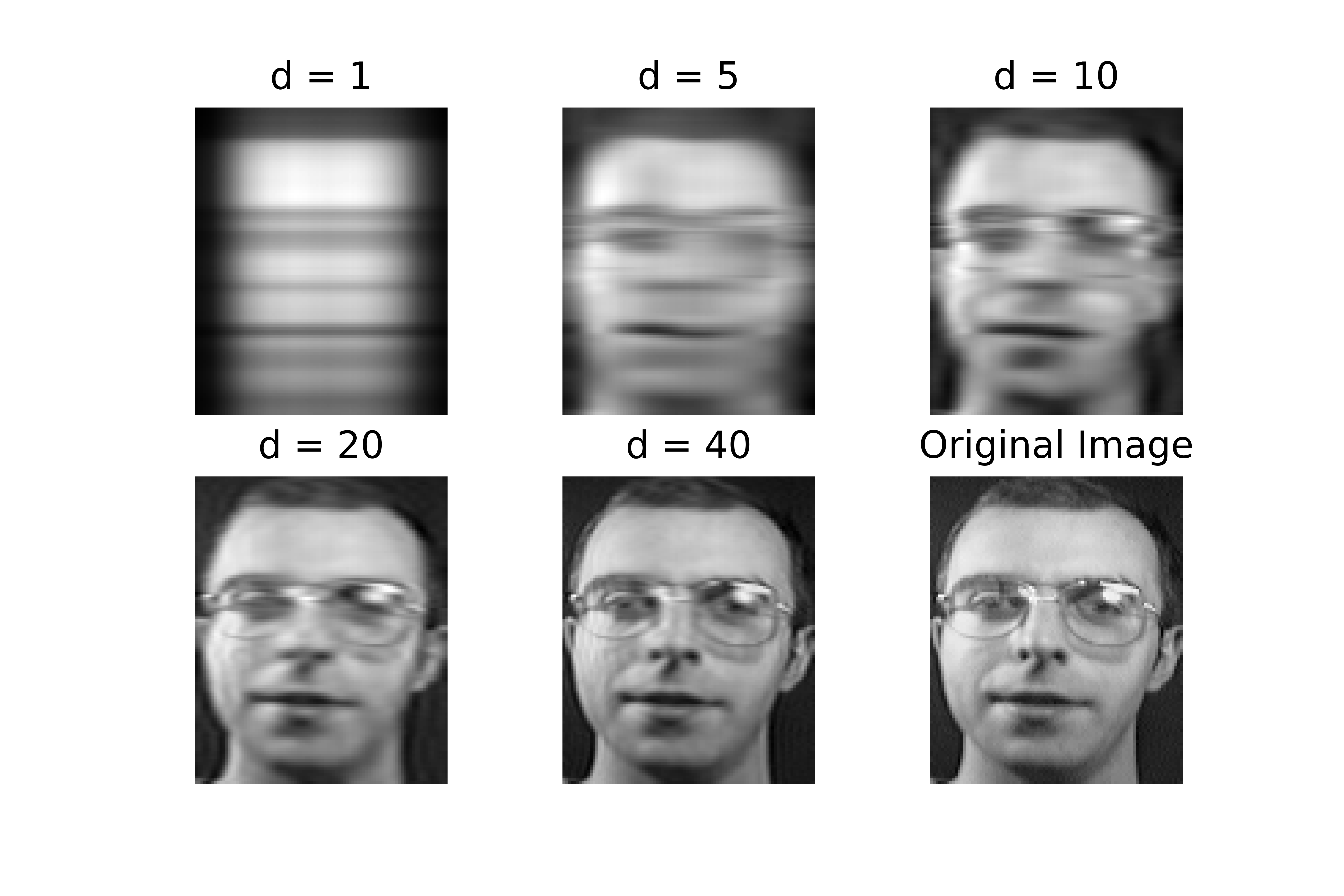}
\vspace{-3em}
\caption{Example ORL image reconstruction by R2DPCA.}
\label{fig:orl_reconstructR}
\end{minipage}%
\begin{minipage}{0.5\textwidth}
\centering
\includegraphics[width=\textwidth,keepaspectratio]{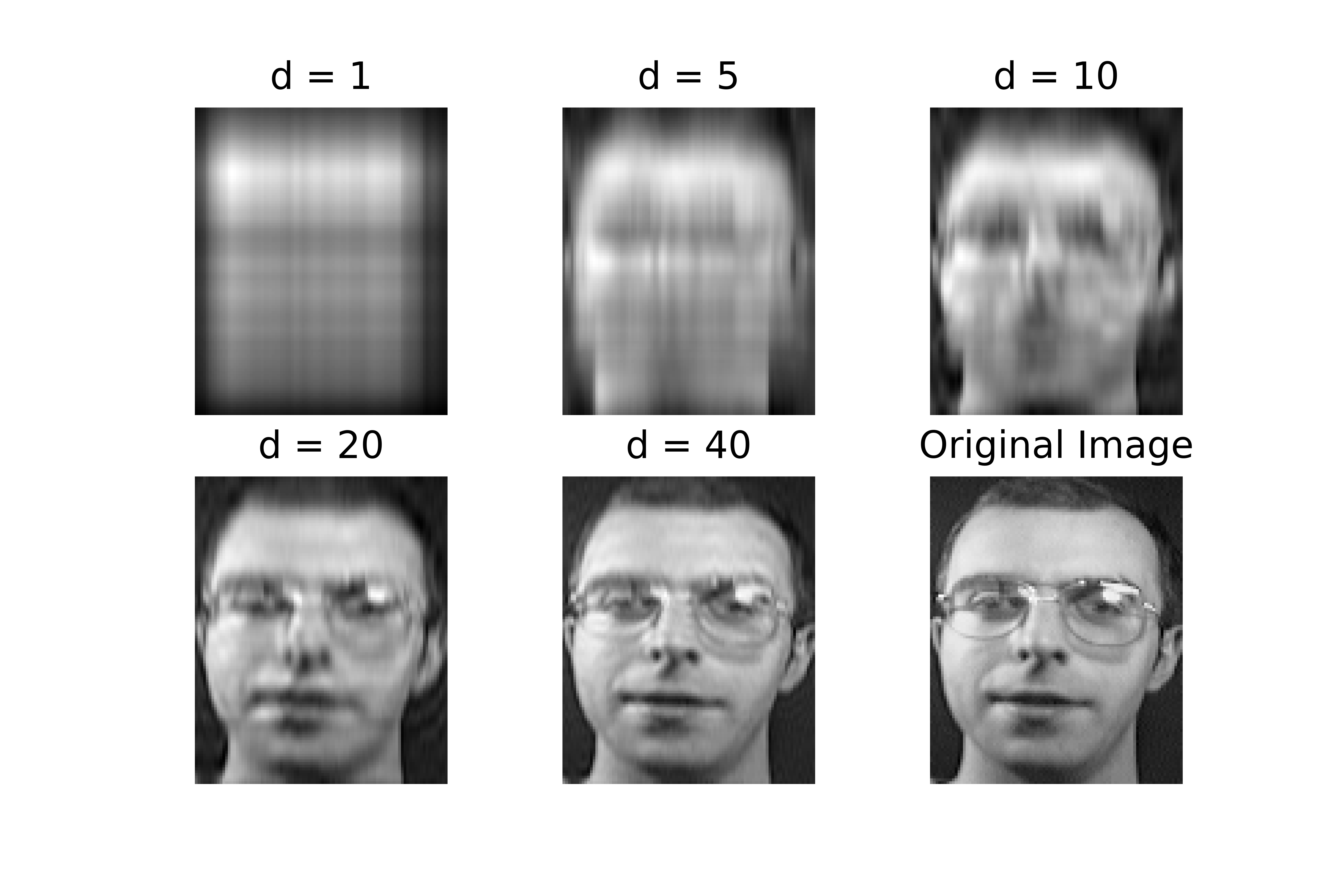}
\vspace{-3em}
\caption{Example ORL image reconstruction by L2DPCA.}
\label{fig:orl_reconstructL}
\end{minipage}
\end{figure*}

Since $\mathbf{U}_1, \ldots, \mathbf{U}_d$ are orthonormal, from (\ref{eq:feature_extraction2}) we can obtain a reconstruction of image $\mathbf{A}_j$:
\begin{equation}
\mathbf{\tilde{A}}_j = \mathbf{U}\mathbf{Y}_j = \sum_{k=1}^d \mathbf{U}_k\mathbf{Y}_{j, k}
\end{equation}
$\mathbf{\tilde{A}}_j$ is of the same dimension as $\mathbf{A}_j$.  If $d = m$, then $\mathbf{\tilde{A}}_j = \mathbf{A}_j$.  Otherwise, $\mathbf{\tilde{A}}_j$ is an approximation for $\mathbf{A}_j$.  See Fig. \ref{fig:orl_reconstructL}.

\subsection{Bilateral 2DPCA}

One major limitation of R2DPCA and L2DPCA is that they each only consider information from either the rows or the columns of an image, but not both.  Another drawback is that they require many coefficients for image representation.  In R2DPCA, an $m \times n$ image can only be reduced to $m \times d$, whereas L2DPCA can only reduce the same image to $d \times n$.  Bilateral 2DPCA (B2DPCA) as proposed in \cite{g2dpca} addresses both of these problems.  It incorporates both row and column information from the images, and is able to reduce an $m \times n$ image to $d_1 \times d_2$, making it more computationally efficient.  \\

\subsubsection{Algorithm}

Given $m \times d_1$ feature matrix $\mathbf{U}$ from L2DPCA and $n \times d_2$ feature matrix $\mathbf{V}$ from R2DPCA, project image $\mathbf{A}_j$ by the following transformation:
\begin{equation}
\mathbf{Y}_j = \mathbf{U}^T\mathbf{A}_j\mathbf{V}
\end{equation}
where $\mathbf{Y}_j$ is of dimension $d_1 \times d_2$.  Similar to conventional 2DPCA, classification is done by KNN and a chosen distance metric.  Experiments in \cite{g2dpca} demonstrate the accuracy and efficiency of B2DPCA.

\section{TWO-DIMENSIONAL LINEAR DISCRIMINANT ANALYSIS}

\subsection{Introduction}

Conventional linear discriminant analysis (LDA) is a popular technique for feature extraction and dimension reduction.  LDA seeks an optimal projection of the data so that variance between classes is maximized while the variance within classes is minimized.  Because it is a supervised method, it often has advantages over PCA, especially in face recognition problems.  When using LDA for face recognition, typically the two-dimensional image matrices must first be converted to one-dimensional vectors.  This causes the between-class and within-class scatter matrices to be of high dimension, making them difficult to accurately calculate.  Furthermore, LDA requires that at least one of the matrices be invertible.  However, they are both high-dimensional, and in practice the number of samples is relatively small.  This all but guarantees that both matrices will be singular.  This is known as the small sample size (SSS) problem.  

An extension of 2DPCA, two-dimensional LDA (2DLDA) as proposed in \cite{2dlda}, avoids the SSS problem.  The between-class and within-class scatter matrices are calculated directly from the original image matrices.  Thus, their dimension is much smaller, making them easy to compute accurately.  In practice one has enough data to guarantee that they are not singular, and it is more computationally efficient to find the desired eigenvectors.

\subsection{Right 2DLDA}

As with R2DPCA, Right 2DLDA (R2DLDA) is equivalent to conventional 2DLDA.  We refer to it as R2DLDA to distinguish it from other generalizations of 2DLDA.\\

\subsubsection{Algorithm}

Let $\mathbf{A}$ be a collection of $M$ image matrices of dimension $m \times n$, where $\mathbf{A}_j$ represents the $j$th image matrix, $j = 1, \ldots, M$.  Each image belongs to one of $C$ classes, where the $i$th class $C_i$ has $n_i$ samples and $\sum_{i=1}^C n_i = M$.  R2DLDA transforms all images $\mathbf{A}$ by a set of discriminating vectors $\mathbf{X} = [\mathbf{X}_1, \ldots, \mathbf{X}_d]$ resulting in projected image matrices
\begin{equation}
\mathbf{Y}_j = \mathbf{A}_j\mathbf{X}, \qquad j = 1, \ldots, M.
\end{equation}
$\mathbf{X}$ is $n \times d$ and its columns are chosen to maximize the 2D Fisher criterion
\begin{equation} \label{J(X)}
J(\mathbf{X}) = \frac{\mathbf{X}^T\mathbf{S}_b^r\mathbf{X}}{\mathbf{X}^T\mathbf{S}_w^r\mathbf{X}}
\end{equation}
where $\mathbf{S}_b^r$ and $\mathbf{S}_w^r$ represent the $n \times n$ between-class and within-class scatter matrices of Right 2DLDA, respectively.  Let $\mathbf{\bar{A}}_i$ denote the average image of the $i$th class, and $\mathbf{\bar{A}}$ the average image of all images.  It follows that
\begin{equation}
\mathbf{S}_b^r = \frac{1}{M} \sum_{i=1}^C n_i (\mathbf{\bar{A}}_i - \mathbf{\bar{A}})^T(\mathbf{\bar{A}}_i - \mathbf{\bar{A}}),
\end{equation}
and 
\begin{equation}
\mathbf{S}_w^r = \frac{1}{M} \sum_{i=1}^C \sum_{j \in C_i} (\mathbf{A}_j - \mathbf{\bar{A}}_i)^T(\mathbf{A}_j - \mathbf{\bar{A}}_i).
\end{equation}
It has been shown that the vectors which maximize (\ref{J(X)}) are the eigenvectors of $(\mathbf{S}_w^r)^{-1}\mathbf{S}_b^r$ corresponding to the $d$ largest eigenvalues.  \\

\subsubsection{Feature Extraction}

The optimal projection vectors can be used for feature extraction.  Let $\mathbf{X} = [\mathbf{X}_1, \ldots, \mathbf{X}_d]$ be an $n \times d$ matrix.  Then for a given image $\mathbf{A}_j$, let
\begin{equation} \label{eq:feature_extraction3}
\mathbf{Y}_{j,k} = \mathbf{A}_j\mathbf{X}_k, \qquad k = 1, \ldots, d.
\end{equation}
The set of projected feature vectors $\mathbf{Y}_{j,1}, \ldots, \mathbf{Y}_{j,d}$ are called the Right Fisher feature vectors of the image $\mathbf{A}_j$.  The Right Fisher feature vectors can be used to form an $m \times d$ Fisher feature matrix $\mathbf{Y}_j = [\mathbf{Y}_{j,1}, \ldots, \mathbf{Y}_{j,d}]$.  Classification is done with KNN and a chosen distance metric.


\subsection{Left 2DLDA}

Similar to the analysis of 2DPCA in \cite{g2dpca}, it can be seen that R2DLDA operates on information contained in the rows of image matrices.  There may be different discriminatory information contained in the columns, thus, a natural extension of R2DLDA is Left 2DLDA (L2DLDA).  The framework for L2DLDA is given in \cite{2d2lda}.\\

\subsubsection{Algorithm}

Let $\mathbf{A}$ be a collection of $M$ image matrices of dimension $m \times n$, where $\mathbf{A}_j$ represents the $j$th image matrix, $j = 1, \ldots, M$.  Each image belongs to one of $C$ classes, where the $i$th class $C_i$ has $n_i$ samples $\sum_{i=1}^C n_i = M$.  L2DLDA transforms all images $\mathbf{A}$ by a set of discriminating vectors $\mathbf{Z} = [\mathbf{Z}_1, \ldots, \mathbf{Z}_d]$ resulting in projected image matrices
\begin{equation}
\mathbf{Y}_j = \mathbf{Z}^T\mathbf{A}_j, \qquad j = 1, \ldots, M.
\end{equation}
$\mathbf{Z}$ is $m \times d$ and its columns are chosen to maximize the 2D Fisher criterion
\begin{equation} \label{J(Z)}
J(\mathbf{Z}) = \frac{\mathbf{Z}^T\mathbf{S}_b^l\mathbf{Z}}{\mathbf{Z}^T\mathbf{S}_w^l\mathbf{Z}}
\end{equation}
where $\mathbf{S}_b^l$ and $\mathbf{S}_w^l$ represent the $m \times m$ between-class and within-class scatter matrices of Left 2DLDA, respectively.  Let $\mathbf{\bar{A}}_i$ denote the average image of the $i$th class, and $\mathbf{\bar{A}}$ the average image of all images.  It follows that
\begin{equation}
\mathbf{S}_b^l = \frac{1}{M} \sum_{i=1}^C n_i (\mathbf{\bar{A}}_i - \mathbf{\bar{A}})(\mathbf{\bar{A}}_i - \mathbf{\bar{A}})^T,
\end{equation}
and 
\begin{equation}
\mathbf{S}_w^l = \frac{1}{M} \sum_{i=1}^C \sum_{j \in C_i} (\mathbf{A}_j - \mathbf{\bar{A}}_i)(\mathbf{A}_j - \mathbf{\bar{A}}_i)^T.
\end{equation}
It has been shown that the vectors which maximize (\ref{J(Z)}) are the eigenvectors of $(\mathbf{S}_w^l)^{-1}\mathbf{S}_b^l$ corresponding to the $d$ largest eigenvalues.  \\

\subsubsection{Feature Extraction}

The optimal projection vectors can be used for feature extraction.  Let $\mathbf{Z} = [\mathbf{Z}_1, \ldots, \mathbf{Z}_d]$ be an $m \times d$ matrix.  Then for a given image $\mathbf{A}_j$, let
\begin{equation} \label{eq:feature_extraction4}
\mathbf{Y}_{j,k} = \mathbf{Z}_k\mathbf{A}_j, \qquad k = 1, \ldots, d.
\end{equation}
The set of projected feature vectors $\mathbf{Y}_{j,1}, \ldots, \mathbf{Y}_{j,d}$ are called the Left Fisher feature vectors of the image $\mathbf{A}_j$.  The Left Fisher feature vectors can be used to form a $d \times n$ Fisher feature matrix $\mathbf{Y}_j = [\mathbf{Y}_{j,1}^T, \ldots, \mathbf{Y}_{j,d}^T]^T$.  Classification is done with KNN and a chosen distance metric.


\subsection{Bilateral 2DLDA}

R2DLDA and L2DLDA suffer from the same limitations as R2DPCA and L2DPCA.  Bilateral 2DLDA (B2DLDA) as proposed in \cite{2d2lda} addresses these shortcomings.  It incorporates both row and column information from the images, and is able to reduce an $m \times n$ image to $d_1 \times d_2$, making it more computationally efficient.  \\

\subsubsection{Algorithm}

Given $m \times d_1$ feature matrix $\mathbf{Z}$ from L2DLDA and $n \times d_2$ feature matrix $\mathbf{X}$ from R2DLDA, project image $\mathbf{A}_j$ by the following transformation:
\begin{equation}
\mathbf{Y}_j = \mathbf{Z}^T\mathbf{A}_j\mathbf{X}
\end{equation}
where $\mathbf{Y}_j$ is of dimension $d_1 \times d_2$.  Similar to conventional 2DLDA, classification is done by KNN and a chosen distance metric.  Experiments in \cite{2d2lda} demonstrate the accuracy and efficiency of B2DLDA.

\section{RANDOM SUBSPACE METHOD}

\subsection{Overview}

In ensemble learning one attempts to train a set of diverse classifiers whose individual outputs are combined into one final decision.  If the classifiers are diverse, that is, if they each make different mistakes, then the hope is that through a sensible combination of the classifiers' decisions that those individual errors will be corrected.  Many techniques for training diverse classifiers exist.  Bootstrap aggregating (bagging) is a popular method.  In bagging, each classifier is trained on a random subset of the training data in order to promote model variance.  The random subspace method \cite{randsubspace} is similar to bagging, but instead of training each model on a random subset of the training data, each model is trained on random samples of features instead of the entire feature set.  This causes individual classifiers to not over-focus on features that appear highly predictive in the training set.  It is generally used with decision trees, though it has been applied to other areas as well.

\begin{figure*}
\centering 
\begin{minipage}{0.47\textwidth}
\centering
\includegraphics[width=\textwidth,keepaspectratio]{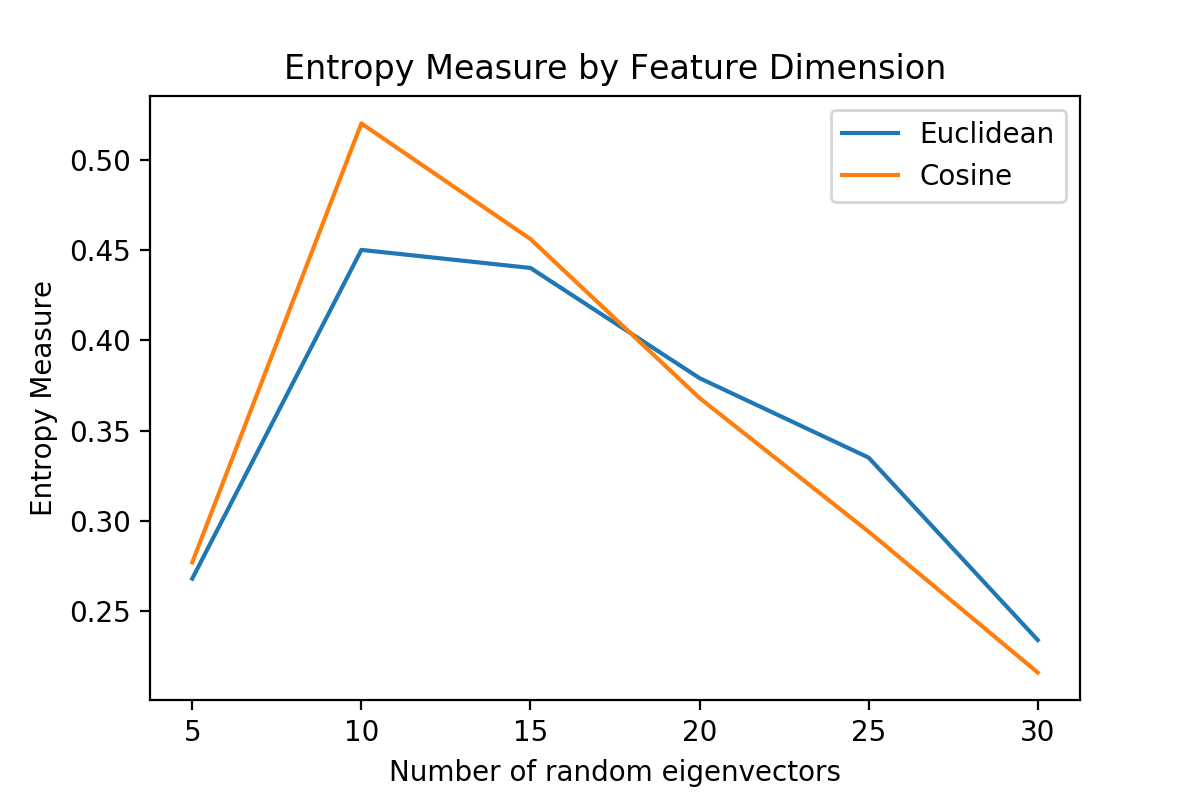}
\caption{Entropy experiment on a subset of MORPH-II with RS-2DLDA.  Selecting 10 random eigenvectors increases entropy for both euclidean and cosine distances.}
\label{fig:entropy}
\end{minipage}%
\begin{minipage}{0.05\textwidth} \
\end{minipage}
\begin{minipage}{0.47\textwidth}
\centering
\includegraphics[width=\textwidth,keepaspectratio]{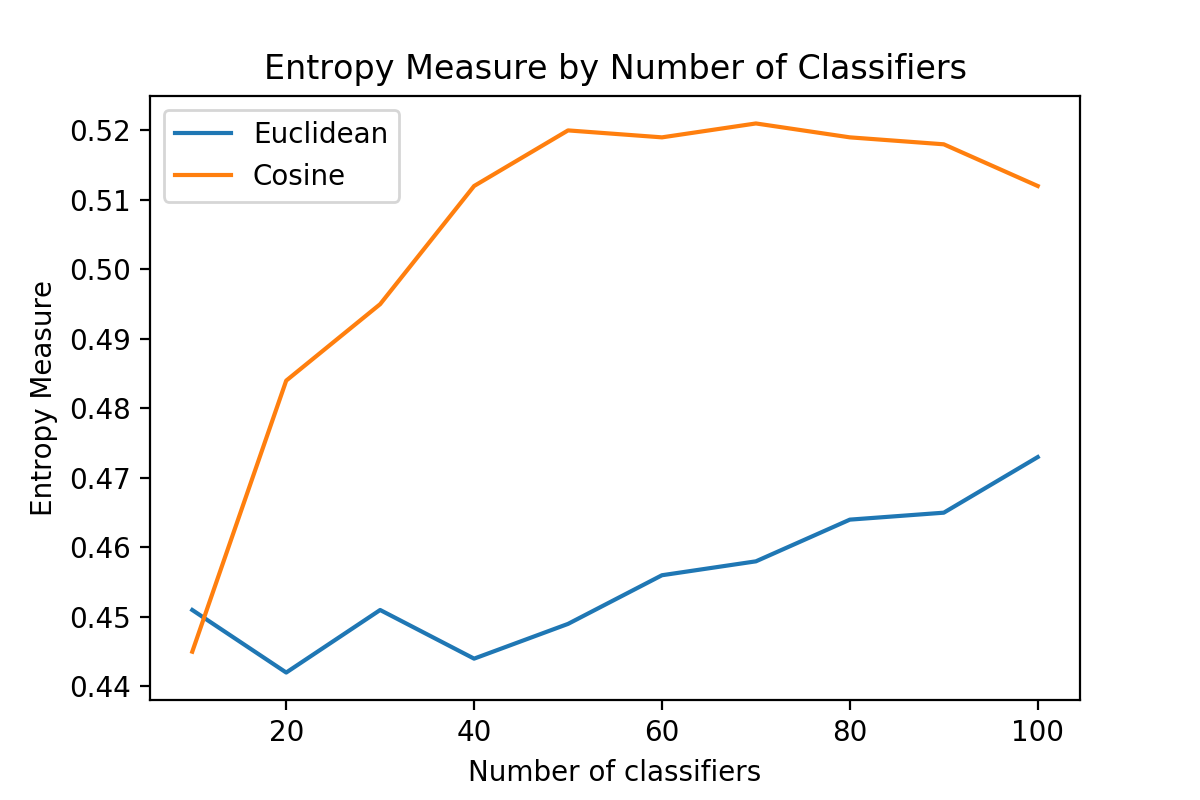}
\caption{Entropy experiment carried out on a subset of MORPH-II with RS-2DLDA.  Train at least 50 random classifiers to increase diversity.}
\label{fig:entropy1}
\end{minipage}
\end{figure*}

\subsection{Application to 2DPCA}

In \cite{rs2dpca}, Nguyen et al. proposed random subspace two-dimensional PCA (RS-2DPCA).  To our knowledge, this is the only application of the random subspace method to any of the two-dimensional variants of PCA, LDA, etc.  In their paper they note that the accuracy of 2DPCA depends heavily on $d$, the number of eigenvectors kept.  Choosing $d$ too low results in poor accuracy, while choosing $d$ large can easily cause overfitting to the training data.  Generally, the $d$ eigenvectors corresponding to the $d$ largest eigenvalues are kept.  However, the eigenvectors that are discarded still contain valuable information.  To overcome these limitations, random samples of the eigenvectors are used to build many classifiers.  As shown in \cite{rs2dpca}, this results in improved and more stable accuracy, and makes it possible to utilize all of the eigenvectors without risk of overfitting.  Unfortunately, the Nguyen et al. approach does not perform well on difficult datasets.  In the next section, we propose our algorithm and demonstrate its improvements over RS-2DPCA.

\section{RANDOM SUBSPACE TWO-DIMENSIONAL LINEAR DISCRIMINANT ANALYSIS}

\subsection{Introduction}

Motivated by RS-2DPCA, we propose random subspace two-dimensional LDA (RS-2DLDA).  Important advantages of RS-2DLDA over RS-2DPCA include:
\begin{itemize}
\item The random subspace is a random sampling of eigenvectors from 2DLDA.  These eigenvectors have been shown to have more discriminative power than those from 2DPCA.
\item Entropy measure is used to select parameters that result in diverse classifiers.
\item Each classifier's reliability is estimated from an adjusted Rand index (ARI) based off of performance on the training data.
\item The ARI scores are used to develop a weighting scheme which is utilized in the final ensemble decision and further boosts accuracy.
\end{itemize}
Although we focus on applying the random subspace method to 2DLDA, it can be easily extended to 2DPCA, because it is merely a random sample of eigenvectors.  We will give analysis of both to illustrate the superiority of 2DLDA over 2DPCA in face recognition.  Note that our application of the random subspace method to 2DPCA is not equivalent to the approach by Nguyen et al. in \cite{rs2dpca}.  They do not consider ways to measure classifier diversity nor does their model incorporate a weighting scheme.

\subsection{Increasing Diversity}

\nocite{ensemble}
The advantage of ensemble systems over single classifiers is that the combination of outputs from many classifiers can often correct for the errors of individual classifiers.  However, this only works when the classifiers are diverse, that is, when each classifier makes different mistakes.  If all classifiers are essentially the same, we cannot hope to correct their individual errors.  There exist many ways to measure classifier diversity.  One such method is entropy measure, which assumes diversity is highest when half of the classifiers are correct for a given test image.  Define $\zeta_i$ as the number of classifiers out of $T$ that misclassify the $i$th image.  Then entropy is defined as 
\begin{equation}
E = \frac{1}{M} \sum_{i=1}^M \frac{1}{T - \lceil T/2 \rceil} \min\{\zeta_i, T - \zeta_i\}
\end{equation}
where $E \in [0,1]$.  Low values indicate similar classifiers, and high values indicate diverse classifiers.  If we can choose parameters that yield a highly diverse set of classifiers, then it is more likely that we will be able to increase the final accuracy with an intelligent combination of the classifier outputs.

We can see in Fig. \ref{fig:entropy} that entropy varies with the number of random eigenvectors we select.  If we don't choose enough eigenvectors, then each classifier is not predictive enough and performs poorly.  On the other hand, if we choose too many, the classifiers are all very similar.  Choosing a moderate value (10 in this case) works well to increase classifier diversity.  

Entropy is also affected by the number of random classifiers we train.  Not training enough results in poor diversity.  Training more classifiers will increase entropy, but to a point.  Fig. \ref{fig:entropy1} illustrates this trend.

\subsection{Estimating Classifier Credibility}

In RS-2DLDA, each classifier is a random sampling of eigenvectors from 2DLDA.  Together these eigenvectors form a matrix.  All images are multiplied by this matrix, which projects them to a new space.  It is impossible to know how well a classifier will perform on the testing data, but we can get an idea based on its performance on the training data.  Since each classifier defines a projection to a new space, we can expect that the classifiers which project images of the same person close to each other but images of different people far apart will perform well on the testing data.  On the other hand, if the projected images of different people are mixed together, and there are no clear boundaries separating the images of one person from the next, then we will expect the classifier to perform poorly.

For a given classifier, take a training image and find its predicted class using KNN on the remainder of the training set.  Do this for all training images, obtaining a prediction for each training image.  Let this set of predictions $\boldsymbol{P}$ define a clustering of the training images.  Let the ground truth values $\boldsymbol{G}$ define another clustering of the training images.  We can expect the classifier to perform well if $\boldsymbol{P}$ is similar to $\boldsymbol{G}$.  That is, we can use a clustering similarity measure to evaluate whether the projection defined by the random sample of eigenvectors appears to preserve or disregard class differences.  The adjusted Rand index \cite{ARI} is one such clustering similarity measure that fits this purpose well.

\begin{figure}
\centering
\includegraphics[width=0.5\textwidth,keepaspectratio]{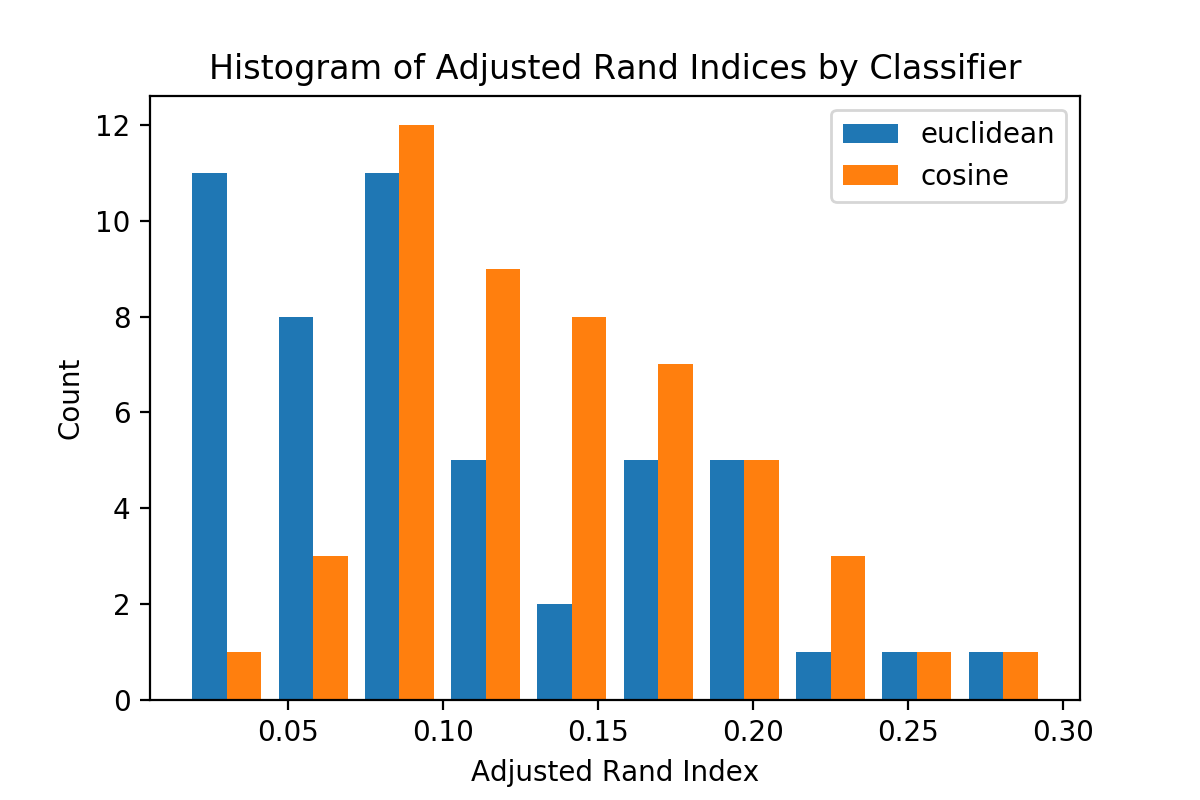}
\caption{Adjusted Rand index experiment performed on a subset of MORPH-II.  On average the ARI are higher for cosine distance than for euclidean.}
\label{fig:ari}
\end{figure}

Given $\mathbf{A}$, a set of $M$ images and two clusterings of these images, namely $\boldsymbol{G} = \{\boldsymbol{G}_1, \ldots, \boldsymbol{G}_C\}$, the ground truth identities of the images and $\boldsymbol{P} = \{\boldsymbol{P}_1, \ldots, \boldsymbol{P}_C\}$, the predicted identities from a given classifier, the overlap between $\boldsymbol{G}$ and $\boldsymbol{P}$ can be summarized in a contingency table $[o_{ij}]$, where each entry $o_{ij}$ is the number of images in common between $\boldsymbol{G}$ and $\boldsymbol{P}$, $o_{ij} = |\boldsymbol{G}_i \cap \boldsymbol{P}_j|$.
\begin{center}
\begin{tabular}{c|cccc|c}
 & $\boldsymbol{P}_1$ & $\boldsymbol{P}_2$ & $\cdots$ & $\boldsymbol{P}_C$ & Sums\\ \hline
 $\boldsymbol{G}_1$ & $o_{11}$ & $o_{12}$ & $\cdots$ & $o_{1C}$ & $a_1$\\
 $\boldsymbol{G}_2$ & $o_{21}$ & $o_{22}$ & $\cdots$ & $o_{2C}$ & $a_2$\\
$\vdots$ & $\vdots$ & $\vdots$ & $\ddots$ & $\vdots$ & $\vdots$ \\
$\boldsymbol{G}_C$ & $o_{C1}$ & $o_{C2}$ & $\cdots$ & $o_{CC}$ & $a_C$\\ \hline
Sums & $b_1$ & $b_2$ & $\cdots$ & $b_C$ & 
\end{tabular}
\end{center}
The adjusted Rand index (ARI) is then calculated as 
\begin{equation}
ARI = \frac{\sum_{ij}\binom{o_{ij}}{2} - [ \sum_i \binom{a_i}{2} \sum_j \binom{b_j}{2} ] / \binom{M}{2}}{\frac{1}{2}[ \sum_i \binom{a_i}{2}+ \sum_j \binom{b_j}{2} ] - [\sum_i \binom{a_i}{2} \sum_j \binom{b_j}{2}] / \binom{M}{2}}.
\end{equation}
ARI ranges from $-1$ to $1$.  High values indicate similar clusterings, while low values mean the clusterings are dissimilar.  The adjusted Rand index is a corrected-for-chance version of the Rand index.  It includes negative values which indicate a Rand index that is less than would be expected if the clusterings were drawn randomly.  

If a given classifier achieves a high ARI, we know in its projected space that images of the same person are clustered close together and images of different people are spread apart.  Thus, we expect these classifiers to outperform those with a low ARI when applied to the testing data.  To take advantage of this, we develop a weighting scheme to give classifiers with a high ARI more influence in the final decision.

In Fig. \ref{fig:ari} one can see the distribution of ARI for a set of 50 random classifiers operating on a subset of MORPH-II.  The ARI are higher on average for cosine distance than for euclidean, suggesting cosine distance may be more suited to face recognition on MORPH-II than euclidean.  The ARI are all positive, indicating that there is some similarity between the classifiers' predictions on the training set and the ground truth values.  Although the ARI are quite low (the highest is less than 0.3) it is important to remember that in ensemble learning we combine many weak classifiers to build one strong classifier.  Indeed, we could have chosen to sample twice as many eigenvectors and easily increased all ARI, but this would come at the expense of classifier diversity.

\subsection{Weighted Majority Voting}

\begin{figure*}
\begin{minipage}{0.47\textwidth}
\centering
\includegraphics[width=\textwidth,keepaspectratio]{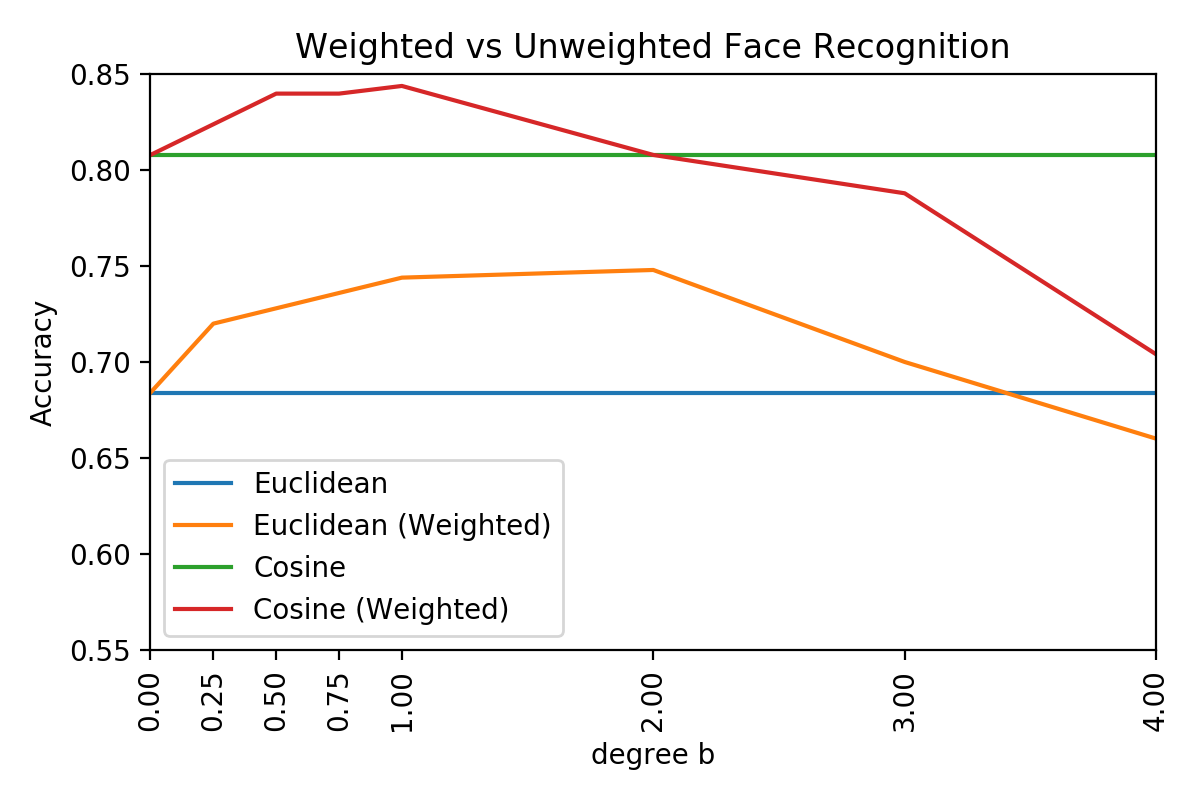}
\caption{Changing the exponent $b$ to which each ARI is raised affects the performance of the weighting scheme.}
\label{fig:b1}
\end{minipage}%
\begin{minipage}{0.05\textwidth}
\
\end{minipage}%
\begin{minipage}{0.47\textwidth}
\centering
\includegraphics[width=\textwidth,keepaspectratio]{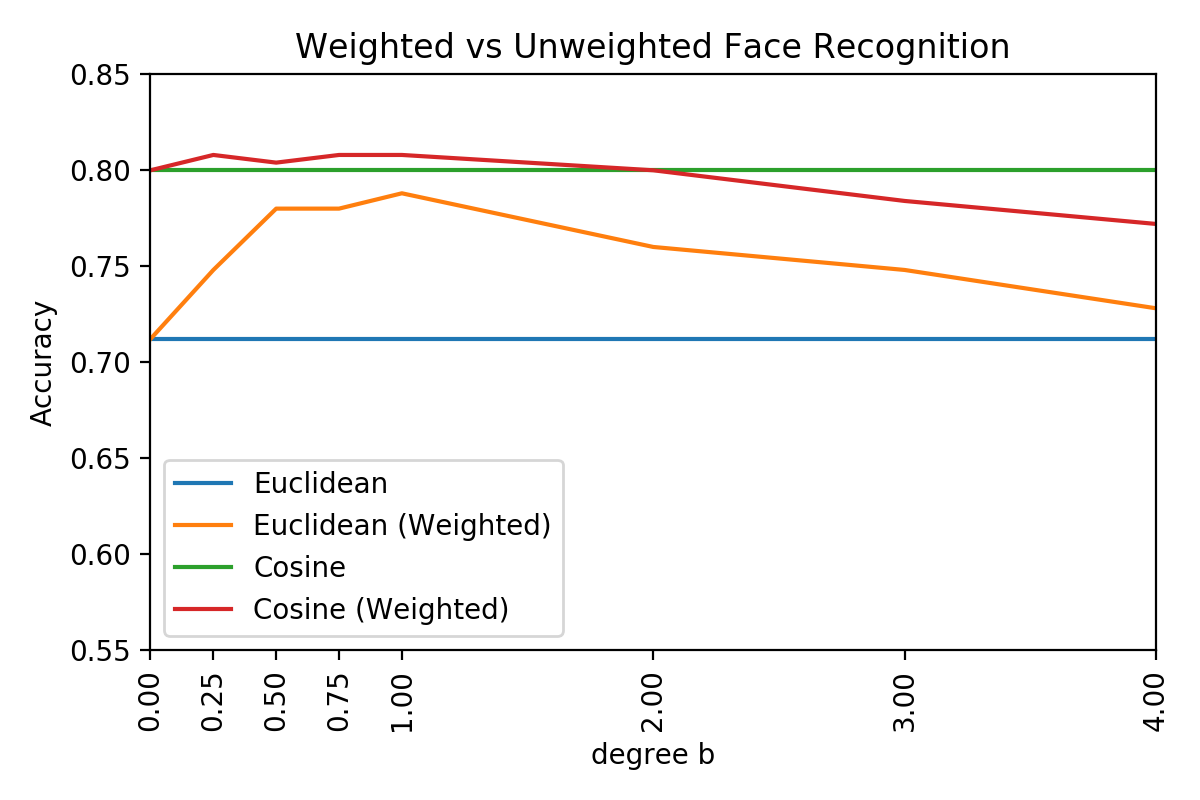}
\caption{There is no value for $b$ that is optimal in every scenario.}
\label{fig:b2}
\end{minipage}
\end{figure*}

In majority voting, each classifier gets one vote.  The final decision is the class with the most votes, regardless of whether the percentage of votes is above $50\%$.  Define the decision of the $t$th classifier as $p_t^{j,c} \in \{0,1\}$, where $t = 1, \ldots, T$, and $c = 1, \ldots, C$.  $T$ is the number of classifiers and $C$ the number of classes.  If the $t$th classifier predicts the $j$th person to belong to class $c$, then $p_t^{j,c} = 1$, and otherwise $0$.  We choose class $c$ if
\begin{equation}
\sum_{t=1}^Tp_t^{j,c} = \max_{c=1}^C\sum_{t=1}^Tp_t^{j,c}.
\end{equation}
If we know that some classifiers are more accurate than others, then we can weight their decisions so that more credible classifiers have a higher influence on the final decision.  This is known as weighted majority voting.  Assign weight $w_t$ to the $t$th classifier.  We choose class $J$ if
\begin{equation}
\sum_{t=1}^Tw_tp_t^{j,c} = \max_{c=1}^C\sum_{t=1}^Tw_tp_t^{j,c}.\\
\end{equation}
We do not require the proportion of support to class $J$ to be over $50\%$.

We expect the classifiers with a high ARI to be more accurate than those with a low ARI.  But how much additional influence should we give to the strong classifiers?  We need a monotone function which maps an ARI to a weight for each classifier.  One simple solution to consider is raising each ARI to a common exponent $b$.  Choosing a low value for $b$ will mean the strong classifiers have marginally more influence than the weak classifiers.  A high value for $b$ would make so that only the strongest classifiers make any real impact on the final decision.  A moderate value for $b$ should give a proper balance and help to increase overall accuracy.

In Fig. \ref{fig:b1} and Fig. \ref{fig:b2} we train 50 random classifiers on a subset of MORPH-II and experiment by varying $b$, the degree to which each ARI is raised.  We can see that a weighing scheme has the potential to substantially boost performance.  However, clearly the best value for $b$ is different in Fig. \ref{fig:b1} than in Fig. \ref{fig:b2}, though they are identical experiments, up to a different initial random seed.  More research needs to be done on how to select an optimal value for $b$.  Other monotone functions, such as the logistic function, could also be considered.

\subsection{Algorithm}

Given the following parameters:

\begin{tabular}{c|p{0.35\textwidth}}
Param. & Description\\ \hline
$\mathbf{A}$ & Set of all images, where $\mathbf{A}_j$ denotes the $j$th image\\
$\{train\}$ & Set of training images\\
$\{test\}$ & Set of testing images\\
$M$ & Total number of images\\
$C$ & Number of people (classes)\\

$\mathbf{S}_w^l, \mathbf{S}_b^l$ & Within-class and between-class covariance matrices from L2DLDA\\
$\mathbf{S}_w^r, \mathbf{S}_b^r$ & Within-class and between-class covariance matrices from R2DLDA\\
$\mathbf{Z}$ & Eigenvectors of $(\mathbf{S}_w^l)^{-1}\mathbf{S}_b^l$\\
$\mathbf{X}$ & Eigenvectors of $(\mathbf{S}_w^r)^{-1}\mathbf{S}_b^r$\\
$d$ & Number of eigenvectors of $\mathbf{Z}$ and $\mathbf{X}$ kept\\
$\mathbf{Y}_j$ & $j$th projected image $\mathbf{Y}_j = \mathbf{Z}^T\mathbf{A}_j\mathbf{X}$\\

$T$ & Number of classifiers\\
$\boldsymbol{G}$ & Ground truth identities for all images\\
$\boldsymbol{P}$ & Predicted training identities, where $\boldsymbol{P}_j$ is the predicted identity of the $j$th image\\
$p_{t}^{j,c}$ & Indicator: 1 if the $t$th classifier predicts the $j$th image to be from the $c$th person, 0 otherwise\\
ARI$_t$ & Adjusted Rand index of the $t$th classifier\\
$b$ & Exponent to which each ARI is raised, resulting in each classifier's weight\\
$w_t$ & Weight given to the $t$th classifier\\
$\boldsymbol{E}$ & Final testing ensemble prediction, where $\boldsymbol{E}_j$ is the predicted identity of the $j$th image
\end{tabular}

and functions:

\begin{tabular}{c|p{0.27\textwidth}}
Function & Description\\ \hline
$knn(img, set, k)$ & Returns the most common class among the $k$ nearest neighbors of $img$ in $set$.\\
$ARI(\boldsymbol{G},\boldsymbol{P})$ & Returns the adjusted Rand index of $\boldsymbol{G}$ and $\boldsymbol{P}$.

\end{tabular}

the algorithm for RS-2DLDA can be summarized as follows:

\begin{algorithm}[H]
\begin{algorithmic}[1]
\small
\FOR[Create $T$ classifiers]{$t=1$ to $T$} 

  \STATE $\mathbf{Z}_t$ = \text{random sample of} $d$ \text{columns of }$\mathbf{Z}$
  \STATE $\mathbf{X}_t$ = \text{random sample of} $d$ \text{columns of }$\mathbf{X}$
  
  \FOR[Project all images]{$j=1$ to $M$}
	\STATE $\mathbf{Y}_j = \mathbf{Z}_t^T\mathbf{A}_j\mathbf{X}_t$
  \ENDFOR
  
  \FOR[Evaluate classifier on training set]{$j$ in $\{train\}$}
  	\STATE $\boldsymbol{P}_j = knn(\mathbf{A}_j, \{train\} \setminus \mathbf{A}_j, k)$
  \ENDFOR  
  
  \STATE $ARI_t = ARI(\boldsymbol{G}, \boldsymbol{P})$
  \STATE $w_t = (ARI_t)^b$
  
  \FOR[Make predictions on testing set]{$j$ in $\{test\}$}
  	\STATE $p_{t}^{j,c} = 0$ for all $c$
  	\STATE $c_{pred} = knn(\mathbf{A}_j, \{train\}, k)$
    \STATE $p_{t}^{j,c_{pred}} = 1$
  \ENDFOR  
\ENDFOR

\FOR{$j$ in $\{test\}$}
  \STATE $\boldsymbol{E}_j = \max_{c=1}^C \Sigma_{t=1}^T w_t p_{t}^{j,c}$
\ENDFOR

\RETURN $accuracy = |\boldsymbol{G} \cap \boldsymbol{E}|$ $/$ $|\{test\}|$

\end{algorithmic}
\label{alg}
\renewcommand{\thealgorithm}{}
\caption{\small RS-2DLDA}
\end{algorithm}

To consider 2DPCA or a different projection scheme (bilateral, left, or right), simply replace the projection in step 5 of the algorithm.  For example, to apply the random subspace method and weighting scheme to L2DPCA, step 5 would become $\mathbf{Y}_j = \mathbf{U}^T\mathbf{A}_j$, whereas for R2DLDA it would be $\mathbf{Y}_j = \mathbf{A}_j\mathbf{X}$.

\section{EXPERIMENTS}

\begin{figure}
\centering
\begin{tabular}{c@{\hskip 4pt}c@{\hskip 4pt}c@{\hskip 4pt}c|c@{\hskip 4pt}c@{\hskip 4pt}c@{\hskip 4pt}c}
\includegraphics[width=0.05\textwidth,keepaspectratio]{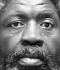}&
\includegraphics[width=0.05\textwidth,keepaspectratio]{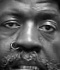}&
\includegraphics[width=0.05\textwidth,keepaspectratio]{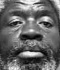}&
\includegraphics[width=0.05\textwidth,keepaspectratio]{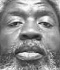}&
\includegraphics[width=0.05\textwidth,keepaspectratio]{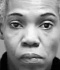}&
\includegraphics[width=0.05\textwidth,keepaspectratio]{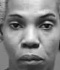}&
\includegraphics[width=0.05\textwidth,keepaspectratio]{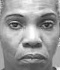}&
\includegraphics[width=0.05\textwidth,keepaspectratio]{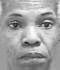}\\ \hline
\raisebox{-.9\height}{\includegraphics[width=0.05\textwidth,keepaspectratio]{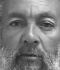}}&
\raisebox{-.9\height}{\includegraphics[width=0.05\textwidth,keepaspectratio]{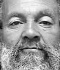}}&
\raisebox{-.9\height}{\includegraphics[width=0.05\textwidth,keepaspectratio]{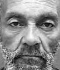}}&
\raisebox{-.9\height}{\includegraphics[width=0.05\textwidth,keepaspectratio]{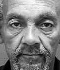}}&
\raisebox{-.9\height}{\includegraphics[width=0.05\textwidth,keepaspectratio]{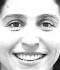}}&
\raisebox{-.9\height}{\includegraphics[width=0.05\textwidth,keepaspectratio]{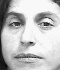}}&
\raisebox{-.9\height}{\includegraphics[width=0.05\textwidth,keepaspectratio]{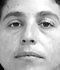}}&
\raisebox{-.9\height}{\includegraphics[width=0.05\textwidth,keepaspectratio]{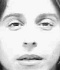}}
\end{tabular}
\caption{Example pre-processed MORPH-II images}
\label{fig:m2_screenshot}
\end{figure}

\subsection{Introduction to the Data}

\subsubsection{MORPH-II}

The MORPH-II dataset \cite{morphii} is a longitudinal dataset collected over five years.  It contains 55,134 images from 13,617 individuals.  Subjects' ages range from 16-77 years of age, and there are an average of four images per person.  MORPH-II is a difficult dataset for face recognition because it suffers from high variability in pose, facial expression, and illumination.  To account for this, all images were pre-processed.  OpenCV was used to automatically detect the face and the eyes in each image.  The images were rotated so that the eyes were horizontal, and then cropped to $70 \times 60$ to reduce noise from the background or the subject's hair.  Finally, all images were histogram equalized with a built-in Python function to help account for the differences in illumination.  

\subsubsection{ORL}
The ORL dataset \cite{orl} contains 40 people each with 10 images of size $112 \times 92$.  There are minor variations in lighting and facial expression, but it is an easy dataset for face recognition.  No pre-processing was done on the images.

\subsection{Experiment Design}

A subset of MORPH-II was used for the experiments.  Among those with 10+ images, 50 arbitrary people were selected.  Five images per person (250 images total) were randomly selected for training, and five for testing (250 images total).  50 different 5-nearest neighbor classifiers were created, each from a random sample of 10 eigenvectors.  The classifiers' predictions on the testing data were combined into one final decision by weighted majority voting.  For ORL, the entire dataset was used.  Five images per person (200 images total) were randomly selected for training, and five for testing (200 images total).  50 different 1-nearest neighbor classifiers were created, each from a random sample of 5 eigenvectors.  The classifiers' predictions on the testing data were combined into one final decision by weighted majority voting.  For completeness, bilateral, right, and left projection schemes of 2DLDA and 2DPCA are considered.  All experiments were repeated thirty times and results averaged to obtain the results in tables \ref{fig:m2_results} and \ref{fig:orl_results}.  Standard error is shown in parentheses.

\begin{table*}[t]
\centering
\normalsize
\begin{tabular}{|*{7}{c|}}
\hline
\multicolumn{7}{|c|}{\textbf{Face Recognition on MORPH-II}}\\ \hline
\multirow{2}{*}{Algorithm} &
\multicolumn{3}{c}{Euclidean} &
\multicolumn{3}{|c|}{Cosine} \\ \cline{2-7}
 & Weighted & Unweighted & Original & Weighted & Unweighted & Original \\ \hline

B2DLDA & .727 (.019) & .678 (.026) & .764 & .781 (.018) & .786 (.015) & .768\\
L2DLDA & .743 (.008) & .735 (.009) & .756 & \textbf{.788} (.010) & .780 (.012) & .776 \\ 
R2DLDA & .704 (.016) & .662 (.018) & .704 & .723 (.018) & .733 (.016) & .704\\
B2DPCA & .706 (.013) & .701 (.013) & .564 & .702 (.018) & .692 (.013) & .556 \\
L2DPCA & .678 (.009) & .667 (.011) & .552 & .670 (.018) & .660 (.016) & .544 \\
R2DPCA & .611 (.010) & .609* (.007) & .580 & .609 (.009) & .612 (.009) & .584\\
 \hline
\end{tabular} 
\caption{Experiments conducted on MORPH-II.  Standard error is shown in parentheses, and top accuracy in bold.  The framework introduced by Nguyen et al. in \cite{rs2dpca} is denoted (*).}
\label{fig:m2_results}
\end{table*}
\
\begin{table*}[t]
\centering
\normalsize
\begin{tabular}{|*{7}{c|}}
\hline 
\multicolumn{7}{|c|}{\textbf{Face Recognition on ORL}} \\ \hline
\multirow{2}{*}{Algorithm} &
\multicolumn{3}{c}{Euclidean} &
\multicolumn{3}{|c|}{Cosine} \\ \cline{2-7}
 & Weighted & Unweighted & Original & Weighted & Unweighted & Original \\ \hline

B2DLDA & .931 (.017) & .924 (.017) & .935 & .939 (.015) & .936 (.016) & .940\\
L2DLDA & .914 (.013) & .909 (.014) & .940 & .937 (.016) & .935 (.017) & .940 \\ 
R2DLDA & .929 (.013) & .923 (.016) & .935 & \textbf{.948} (.015) & .943 (.013) & .945\\
B2DPCA & .914 (.013) & .911 (.014) & .870 & .908 (.015) & .908 (.013) & .865 \\
L2DPCA & .895 (.011) & .893 (.010) & .865 & .884 (.012) & .884 (.013) & .860 \\
R2DPCA & .905 (.010) & .903* (.011) & .895 & .916 (.016) & .914 (.016) & .895\\
 \hline
\end{tabular}
\caption{Experiments conducted on ORL.  Standard error is shown in parentheses, and top accuracy in bold.  The framework introduced by Nguyen et al. in \cite{rs2dpca} is denoted (*).}
\label{fig:orl_results}
\end{table*}

\subsection{Analysis}
From the results in tables \ref{fig:m2_results} and \ref{fig:orl_results}, the difficulty of MORPH-II is apparent.  The highest accuracy achieved, 0.788, was 16\% less than the highest for ORL (0.948).  In the MORPH-II experiments, performance increases substantially when cosine distance is used instead of euclidean.  This is likely due to the fact that MORPH-II suffers from high variability in illumination, whereas ORL does not.  Cosine distance is a measure of similarity, not magnitude, so we see boosted accuracy on MORPH-II, but only minor improvements for ORL. 

In general, the 2DLDA algorithms outperform their 2DPCA counterparts.  This is likely due to the fact that the eigenvectors from 2DLDA have more discriminative power for face recognition than those from 2DPCA.

In general the weighting scheme increases accuracy.  However, in some cases the unweighted algorithm achieves better performance, and in other incidents the original (not random subspace) algorithm is the best.  We can be confident that the random subspace method is in general effective, and that the weighting scheme will in most cases increase accuracy.  More research needs to be done on parameter selection.  One obvious direction for future work is in the selection of $b$, the exponent to which all ARI are raised to determine the weighting scheme.  It is obvious that one choice of $b$ does not generalize well to other algorithms (a value for $b$ that works well with L2DLDA may not generalize well to R2DPCA, for example).  A more systematic and robust way of selecting $b$ (and other parameters) is needed to ensure increased performance regardless of algorithm or dataset.

The Nguyen et al. framework in \cite{rs2dpca} achieves satisfactory performance on ORL, but it performs quite poorly on MORPH-II.  Although the high accuracies achieved on ORL are not replicated on MORPH-II, the contributions presented in this paper significantly increase accuracy.  First, considering the cosine distance metric helped to account for the variable illumination of MORPH-II.  Using the eigenvectors from 2DLDA significantly increased recognition accuracy, and considering multiple projection schemes (bilateral, left, and right) showed that one scheme is not always better than the others.  Finally, the weighting scheme proposed here further boosts accuracy.

\section{CONCLUSIONS}

A novel algorithm for face recognition, RS-2DLDA, is presented and evaluated on MORPH-II and ORL datasets.  It outperforms previously proposed RS-2DPCA \cite{rs2dpca} by utilizing multiple distance metrics and projection schemes.  RS-2DLDA further benefits from a weighting scheme that increases accuracy.  Future work will include investigation into the key differences of the bilateral, left, and right versions of 2DLDA and 2DPCA, and exploration into randomly sampling eigenvectors with replacement.  More challenging face recognition problems will also be considered.  Finally, an optimized weighting scheme will be sought out that is effective regardless of dataset difficulty or algorithm used.  






\section*{ACKNOWLEDGMENT}

This work was conducted at an NSF-sponsored Research Experience for Undergraduates (REU) program at University of North Carolina Wilmington.  I would like to thank Dr. Cuixian Chen, Dr. Yishi Wang, and Troy Kling for their dedication and support.


\bibliographystyle{plain}
\bibliography{bib1.bib}
\end{document}